# GENIE: Generative Note Information Extraction model for structuring EHR data

Huaiyuan Ying[1], Hongyi Yuan[1], Jinsen Lu[2], Zitian Qu[3], Yang Zhao[4], Zhengyun Zhao[1], Isaac Kohane[5], Tianxi Cai[5], Sheng Yu[1]

[1] Department of Statistics and Data Science, Tsinghua University, Beijing, China

[2] Department of Statistics and Data Science, Tsinghua University, Beijing, China

[3] Zhili College, Tsinghua University, Beijing, China

[4] Weiyang College, Tsinghua University, Beijing, China

[5] Department of Biomedical Informatics, Harvard Medical School, Boston, United States of America

Corresponding Author: Sheng Yu, Weiqinglou 209, Department of Statistics and Data Science, Tsinghua University, Beijing, China; Email: syu@tsinghua.edu.cn; Tel: 010-62783842.

# ABSTRACT

Electronic Health Records (EHRs) hold immense potential for advancing healthcare, offering rich, longitudinal data that combines structured information with valuable insights from unstructured clinical notes. However, the unstructured nature of clinical text poses significant challenges for secondary applications. Traditional methods for structuring EHR free-text data, such as rule-based systems and multi-stage pipelines, are often limited by their time-consuming configurations and inability to adapt across clinical notes from diverse healthcare settings. Few systems provide a comprehensive attribute extraction for terminologies. While giant large language models (LLMs) like GPT-4 and LLaMA 405B excel at structuring tasks, they are slow, costly, and impractical for large-scale use. To overcome these limitations, we introduce GENIE, a Generative Note Information Extraction system that leverages LLMs to streamline the structuring of unstructured clinical text into usable data with standardized format. GENIE processes entire paragraphs in a single pass, extracting entities, assertion statuses, locations, modifiers, values, and purposes with high accuracy. Its unified, end-to-end approach simplifies workflows, reduces errors, and eliminates the need for extensive manual intervention. Using a robust data preparation pipeline and fine-tuned small scale LLMs, GENIE achieves competitive performance across multiple information extraction tasks, outperforming traditional tools like cTAKES and MetaMap and can handle extra attributes to be extracted. GENIE strongly enhances real-world applicability and scalability in healthcare systems. By open-sourcing the model and test data, we aim to encourage collaboration and drive further advancements in EHR structurization.

# Introduction

Electronic health records (EHRs) have been widely implemented in many countries and have facilitated clinical workflow, billing, and data access for both care providers and patients. Beyond their primary purpose, EHRs can facilitate critical downstream goals such as supporting clinical decision-making through machine learning algorithms, enabling data collection for research, informing government administration, and guiding policy design, thereby offering immense value to healthcare [1-3]. EHR data can be broadly categorized into structured data (coded or uncoded) and unstructured narrative data, with a significant usability gap between the two formats due to the complexities and challenges of processing narrative data. Structured EHR data can be efficiently queried from databases and readily transformed into research variables, machine learning features, or statistical metrics, making it highly versatile for analysis [4]. In contrast, narrative data typically require advanced natural language processing (NLP) techniques for analysis. However, they offer far more detailed and comprehensive information than structured data, making them invaluable for downstream analyses [5]. For example, signs and symptoms as well as treatment side effects are typically only documented in clinical notes. Extracting such information in a standardized format can greatly enhance researchers' ability to effectively incorporate information from these variables.

Structuring narrative EHR data, henceforth referred to as EHR structurization for brevity, is a highly specialized branch of NLP. Currently, its implementation in hospitals is highly limited and typically tailored to specific research projects. Here, we briefly review EHR structurization to understand its tasks and challenges. EHR structurization is primarily focused around identifying medical terms and analyzing their semantics and attributes, with the following common modules:

- Medical term recognition: after preprocessing steps such as format cleaning, medical term

recognition is often the first analysis to perform, which can be achieved either by string matching or by named entity recognition (NER) techniques. String matching is typically implemented as trie search, such as the Aho-Corasick algorithm[6]. These search algorithms are typically implemented against a very large lexicon such as the Unified Medical Language System (UMLS)[7] and the Biomedical Informatics Ontology System (BIOS)[8]. NER can be achieved by statistical NLP such as noun phrase parsing[9,10], or by machine learning (ML) models for NER as a token-level classification task[11].

- Entity linking or term standardization: to facilitate feature extraction, information retrieval, and interoperability, the identified terms, which can be potentially nonstandard or even ambiguous, need to be standardized or coded. Entity linking resolves the identified terms to well defined concepts in ontologies. Commonly used ontologies for entity linking include the UMLS, Medical Subject Headings (MeSH)[1], SNOMED CT [12], International Classification of Diseases (ICD) [13], RxNorm [14], Current Procedural Terminology (CPT) [15], etc. The result of NER can be mapped to the target ontologies via different approaches, including exact match against the target terminology[16], linguistic search[15], or embedding-based search[17]. EHR notes contain a large number of abbreviations, which cannot be automatically linked to target ontology, and the same abbreviation can often refer to multiple concepts. For example, RA can refer to Room Air, Rheumatoid Arthritis, Refractory Anemia, Right Atrium, etc. The meaning of the abbreviations need to be resolved together with the context, which is called word sense disambiguation (WSD). WSD can be implemented as part of entity linking or as a separate module by itself. Term standardization shares the same purpose as entity linking, with the former outputs standardized

---

[1] https://www.nlm.nih.gov/mesh/meshhome.html

terms, and the latter outputs concept codes.

- Assertion analysis: as mentions of medical terms in EHRs may have different implications, the mentions need to be classified along the axes of presence (e.g., present, absent, or possible), temporality (e.g., current or past medical history), experiencer (e.g., patient or family history), etc., which are collectively called assertion analysis. Early systems mainly used rule-based methods for assertion analysis. For example, NegEx[18] and ConText[19] used regular expressions for the classification, with the former classified only absent versus not absent, and the latter extended the former to 5 labels. NILE[20] used a sequence of parsers to achieve classifications of present, absent, possible, family history, or ignore (term not implicating presence and should be ignored in the output). The 2010 i2b2/VA NLP challenge provided annotated data for 6 labels: present, absent, possible, hypothetical, conditional, and someone else[21], which facilitated the development of many ML-based models afterwards. The increasingly detailed classification can be beneficial for downstream analysis.

- Location and modification: the body location of conditions and procedures is one of the most clinically important attributes. Locations are also associated with assertion analysis, as absence of problems in one area does not imply absence of problems overall. Location analysis can be achieved by rules, dependency parsing, or ML-based relation analysis [22-24]. However, locations which appear at a distant sentence from its subject term can be difficult to identify by conventional methods. Modification analysis aims to attach clinically informative modifiers to the identified terms, such as severity, chronicity, side, color, and texture. It is similar to and easier than location analysis and can also be achieved using rules or dependency parsing.

- Values and units: values, such as clinical measurements and attributes, laboratory test results, and

medication dosages, are important information for medical studies, and there are many values that are not available as structured data and need to be extracted from the EHR text. Specific values can be easily extracted by regular expressions[25,26]. However, it is challenging to devise a generic text pattern for values not following the basic subject-value-unit format. Moreover, values in EHR notes may also require special inference for the omitted subjects (e.g., vital signs and test panels) and units.

Implementing an EHR structurization system requires the integration of numerous complex modules, as highlighted above. However, few existing systems incorporate all these modules. For instance, MetaMap [9] supports only NER and entity linking, while cTAKES [16] and CLAMP [17] extend this to include assertion analysis. NILE [20] further integrates location and modification analyses. The interdependencies among these modules pose significant challenges, particularly for system maintenance, as upgrading a single module can risk destabilizing the entire system. These practical challenges severely limit the implementation of EHR structurization systems in hospital settings.

Over the past two years, we have witnessed remarkable advancements in large language models (LLMs), which demonstrate vast knowledge gained from pretraining on massive and diverse corpora, including biomedical texts [27,28]. The rapid progress has resulted in increasingly powerful LLMs capable of processing much longer inputs (commonly at least 8K tokens), making them well-suited for tasks like processing and structuring EHRs. Using one same model to carry out different tasks is particularly attractive, because theoretically one can replace the many modules of the EHR structurization system with a single model, which will significantly simplify the system maintenance and upgrade, as well as providing significantly more intelligent text analysis than conventional models. However, even using proprietary models, such as the GPT-4o, crafting an instruction prompt for each task that stably produces satisfying results is difficult. More importantly, LLMs are notably slow and costly. Calling LLMs task by task, each with lengthy instructions, makes structurizing a single EHR note take well over 1 minute and unaffordable at large scale. In addition, EHR processing must comply with privacy protection rules, and not many institutions have private infrastructures to access proprietary LLMs for EHR structurization.

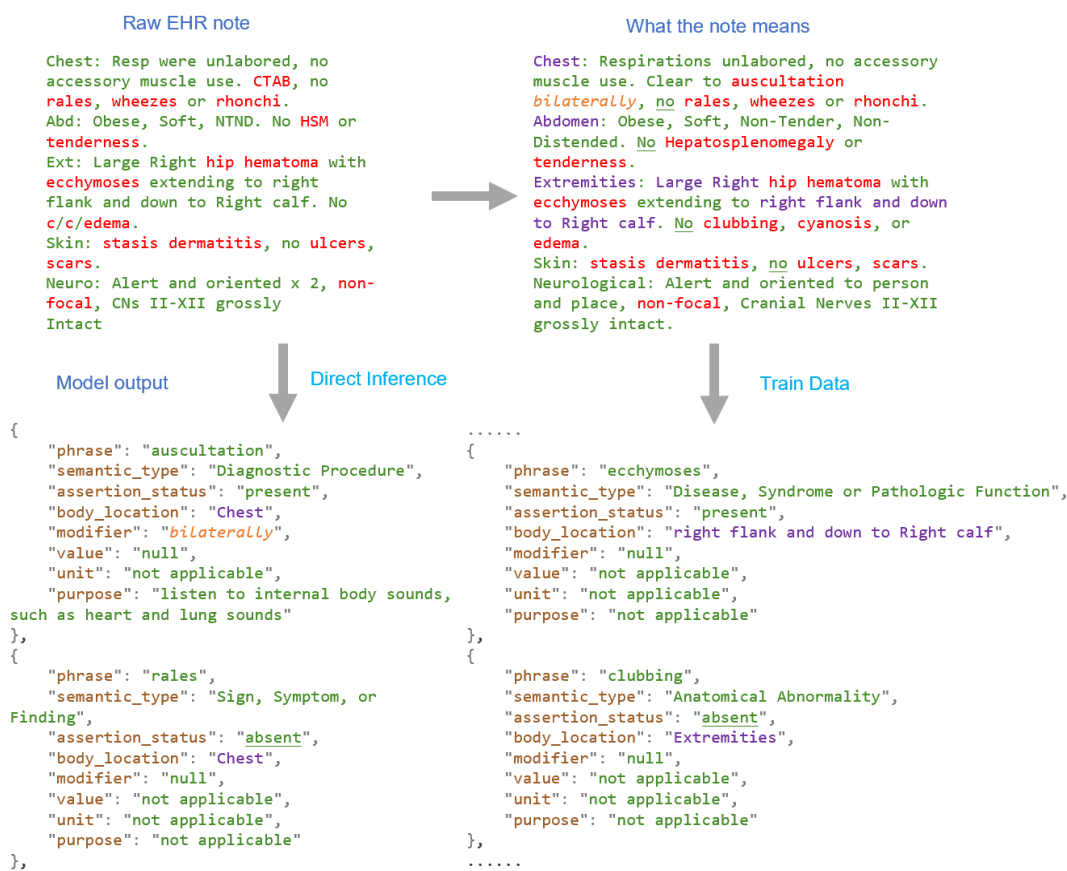

**Figure 1.** An example of our model rationale. Typical raw EHR notes contains plenty of jargons and acronyms, which can be restored for annotation process. When processing training data, we extract attributes and full-name phrases from the restored version, but the input will keep the raw version. The training process will lead our model to learn extract the understandable phrase, semantic types, assertion status, locations, modifiers, values, units and purposes from raw EHR data and output a json-format structured results.

In this work, we propose **GENIE**, an open source **Ge**nerative **N**ote **I**nformation **E**xtraction system for end-to-end EHR structurization. The English version of GENIE is fine-tuned from Llama-3.1-8B-Instruct. As shown in Figure 1, It performs NER and generates each term's text (with automatic standardization for abbreviations), semantic type, assertion status, locations, modifiers, values, units, and purposes as JSON format in a single pass, thereby achieving the following goals simultaneously: 1. Simplify the complex EHR structurization system as a single model; 2. Single pass processing without instruction to significantly reduce the processing time; 3. Run locally on consumer level hardware (e.g., the NVIDIA RTX 3090 with 24GB VRAM). In this paper, we introduce the training pipeline for GENIE and outline how individual task labels are annotated with the assistance of general-domain LLMs and how these tasks are integrated. We also provide an expert annotated EHR dataset and systematically evaluate the performance on each analysis. Experimental results demonstrate the model's efficiency and accuracy.

## Methods

In this section, we introduce the pipeline for training GENIE. We use the discharge reports of MIMIC-III[29] as the EHR corpus, under the Azure license for using GPT to facilitate data annotation. The input to GENIE are whole EHR notes or sections of them, if they are too long. The output are identified medical terms and their attributes by the order they appear in the notes. Below, we first introduce the annotation of each attribute, then the aggregation of the data and the model training as in figure 2.

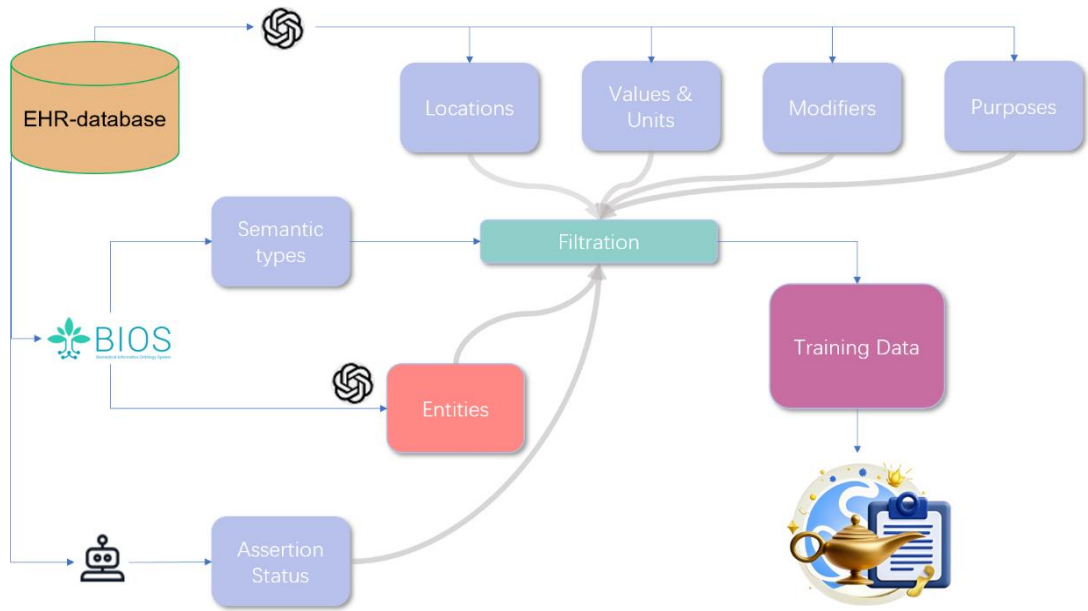

**Figure 2.** The data integration pipeline of our GENIE model. We first extract the phrase and corresponding semantic types using Forward Maximum Matching and BIOS ontology with slight assistance from GPT on acronyms. An independent LLM is trained for detecting assertion status, while other attributes are mainly extracted from GPT. All the attributes are then integrated into entity-wise results, filtered by semantic types, to form the training data.

**EHR preprocessing and term recognition**

We use the forward maximum matching method (i.e., trie search) with BIOS v3 as the terminology for term annotation. Although BIOS provides a very comprehensive coverage of clinical terms with 36 million terms in English, the MIMIC notes need to be preprocessed for trie search to work effectively. First, MIMIC notes contain a large number of line breaks for formatting purposes that break sentences. Those line breaks will disrupt term matching and assertion analysis and should be removed, but the true line breaks (ends of paragraphs or list items) should be kept. Second, the terms in BIOS are identified from research papers and may miss certain abbreviations that appear only in EHRs. In addition, the abbreviations in EHRs require WSD with context, which cannot be achieved by trie search alone.

Therefore, we instruct ChatGPT to perform line break restoration and abbreviation expansion to facilitate term recognition. The prompts are provided in Supplementary Materials.

We then use trie search to identify BIOS terms in the preprocessed notes. With raw EHR as the input and expanded abbreviations in the annotated output, GENIE will learn WSD automatically. The semantic types of the identified terms in BIOS also serve as an output attribute. The output terms are filtered and limited to the following semantic types: Anatomical Abnormality, Cell or Molecular Dysfunction, Chemical or Drug, Clinical Attribute, Diagnostic Procedure, Disease, Syndrome or Pathologic Function, Eukaryote, Individual Behavior, Injury or Poisoning, Laboratory Procedure, Mental or Behavioral Dysfunction, Microorganism, Neoplastic Process, Physiology, Sign, Symptom, or Finding, and Therapeutic or Preventive Procedure. Terms of other semantic types are considered nonessential in EHR structurization or they may be attributes instead of base terms (e.g., body locations).

**Assertion status**

We create an assertion status annotation model based on the 2010 i2b2/VA NLP challenge to annotate the identified terms in the MIMIC data. The i2b2/VA data provides annotations of six categories: Present, Absent, Possible, Conditional, Hypothetical, and Not_associated_with_the_patient. The annotation model is fine-tuned based on Llama-2-7b. The input consists of an EHR note with target terms denoted by double brackets, and the output lists the terms and their classifications following the format: "term: status\n". The trained model achieves 97% accuracy on the i2b2/VA test data.

We apply the annotation model to the terms identified in the previous step. Although model's the test accuracy exceeds all existing models, by observation we find the annotation accuracy of Conditional is not satisfying, due to the rare occurrence of the label in the i2b2/VA data. Thus, we convert the assertion status of all the terms following the word 'allergy' to Conditional by rule. Additionally, we introduce a new category and convert the status of the terms which contain capital letters and are immediately followed by ':' as "Title".

**Locations, modifiers, values and units, and purposes**

The annotation processes for these attributes follow a similar methodology. As no publicly available datasets exist for these tasks, the annotation of these attributes are performed by GPT. The exact GPT model is chosen to be the most affordable one with acceptable accuracy at the time of the annotation and ranges from ChatGPT to GPT-4o. Through iterative prompt engineering, we find the following prompt effective and stable at minimizing missed terms and mismatched results (using location as an example):

*"Here is one part of a medical note:\n" + paragraph + "\nFor each of the entities from the note, which are marked with double curly braces in the note, output its body location according to the note (Format: entity: location), which should be an exact substring from the note, and the location should be a noun or noun phrase or direction. If the note did not specify the location explicitly, only print 'null'.\nHere are the terms:" + "\n".join(terms)*

The attributes are extracted independently. Apparently, not all attributes are applicable to every semantic type. For instance, a disease or symptom should not have a "purpose" attribute. To address this, we construct a table that specifies the applicable attributes of each semantic type. During post-processing, if a term should not have a specific attribute, it is marked as "not applicable" to differentiate it from "null", which indicates that the EHR paragraph genuinely lacks the relevant information.

**Data integration and training**

We integrate the identified terms with their corresponding attributes to form the training data. The matching process is guided by term order and relative position in the note in order to find which term in the recognized term list should be matched when there are multiple occurrences of the term. Since most of the attributes are generated by LLMs, it is possible that the values of an attribute cannot map one-to-one back to the identified terms of the note due to generation errors. If an attribute cannot be associated with any term, the entry is dropped and recorded, and notes with excessive mismatches are discarded.

After the integration, the output consists of a list of recognized terms with all their associated attributes, formatted into JSON for further processing. As GPT preprocessing is unlikely to be available in most EHR structurization scenarios, we use the raw EHR as the input when training GENIE. To balance the maximum input length and output quality, we retain only samples with fewer than 8,000 tokens. As the text length ratio between input and output is roughly 1:8, we recommend section EHR notes to pieces shorter than 800 tokens to make sure the output is complete.

All training procedures are conducted on 8 NVIDIA A800 80G GPUs. The assertion status model, based on Llama-2-7b, is trained on 1,200 samples for 6 epochs. The GENIE model, based on Llama-3.1-8b, is trained on 180,000 samples for 3 epochs. The training employs the cross entropy loss function with a learning rate of 2e-5.

# RESULTS

*Test Datasets*

We randomly selected 964 notes from the MIMIC-III dataset to serve as the development set, which was further divided into 2,000 paragraphs of suitable length. These notes were excluded from the training data. Among the development set, 24 paragraphs were manually annotated by two expert annotators, with each paragraph assigned to one annotator. The resulting gold-standard annotations contain a total of 448 phrases: 235 phrases include body locations, 128 phrases have modifiers, and 144 phrases include values and units. One pair of acronym and full name will be viewed as one phrase. The correctness of the *purpose* attribute was not numerically evaluated, as its narrative nature makes it unsuitable for the verification process.

The following experiments and results are conducted on the test set. For evaluation metrics, we use simple accuracy for *assertion status* and other attributes, and F1-score for the phrases. To address cases where predicted attributes differ in expression (e.g., units) from the gold standard, GPT-4o is used to assess equivalence. However, for phrase extraction, the F1-score requires an exact match, distinguishing it from token-level F1 evaluation. Considering that some methods map abbreviations to their full names, whether predicting an acronym or the full name will be judged as a correct match during computation of F1-scores.

*Baseline Models*

To the best of our knowledge, few models address the structuring task in the same manner as our approach. We consider the most widely used methods and tools for comparison, specifying the attributes they are capable of extracting:

cTAKEs [16]: Apache cTAKES is an open-source natural language processing system designed to extract information from clinical free-text in electronic medical records. It identifies codable entities, events, semantic types (using UMLS), time stamps, and limited assertion statuses. However, in the clinical default pipeline they provide, we can only extract assertion and phrases.

CLAMP [17]: The Clinical Language Annotation, Modeling, and Processing (CLAMP) Toolkit, developed at the University of Texas Health Science Center at Houston, comprehensively recognizes and encodes clinical information in narrative patient reports. It supports the extraction of entities, locations, semantic types, and assertion statuses. However, the download of CLAMP has deprecated, so we cannot evaluate the performance.

METAMAP [9]: MetaMap, developed by the National Library of Medicine (NLM), is a tool for identifying UMLS concepts in texts. It primarily focuses on entity recognition and semantic type extraction, including the identification of acronyms.

Additionally, there are similar works that focus on extracting specific attributes, such as EXTEND [25] and NILE [20]. However, these tools require users to predefine the attributes to be extracted, rather than performing fully automated processing. For this reason, we exclude them from our baseline comparison, as our structuring process does not involve pre-specification of attribute ranges.

Table 1. Main results on the human-labelled test sets, using Accuracy (Acc) or F1-value (F1). If one method cannot extract the corresponding attribute, we leave the blank with '-'.

| Attributes | Phrase | Location | Modifier | Value | Unit | Status |
|---|---|---|---|---|---|---|
| Metric | F1 | Acc | Acc | Acc | Acc | Acc |
| **GENIE** | 0.837 | 0.867 | 0.934 | 0.820 | 0.812 | 0.912 |
| **cTAKES** | 0.182 | - | - | - | - | 0.748 |
| **MetaMap** | 0.172 | - | - | - | - | |

The results are presented in Table 1, which demonstrates that our model achieves competitive performance across all attribute extraction tasks, surpassing other existing methods. Notably, the GENIE model is capable of extracting several attributes that other software cannot handle.

Two points in Table 1 warrant further discussion. First, the reported metrics may be slightly underestimated. We observed instances where equivalent results were not recognized by the GPT evaluation process. Second, the assertion status accuracy is lower compared to our single-task LLM model. While it is common for comprehensive models to sacrifice some performance on specific tasks, particularly under long-context inference conditions, we believe the GENIE model exhibits a bias toward predicting *"present"*. This issue may stem from improper post-processing of mismatched status results or existing error patterns in the original i2b2 database. A future update to GENIE is expected to address these limitations and further improve performance.

# DISCUSSION

*Case Study*

To further analyze the rationale behind our data preparation pipeline and evaluate the efficacy of the GENIE model, we manually inspected selected test samples. All input paragraphs are real cases from the MIMIC-III dataset, and the complete outputs cannot be provided due the license of MIMIC.

Our GENIE model demonstrates a robust ability to recognize and restore abbreviated patterns in EHRs, including acronyms, irregular formats, and generic terms. For example, the sentence:

**"Chem    140, 3.6, 103, 26, 20, 1.1 and 107."**

is accurately expanded into seven entries, assigning each value to the corresponding subject (e.g., sodium, potassium) and inferring appropriate units.

Additionally, the model restores abbreviations such as: Recognize **"HSM"** as **"hepatosplenomegaly"**, "c/c/e" as **"clubbing, cyanosis, and edema"** and output the three terms one by one, and **"NPH"** for **"neutral protamine hagedorn"**. Numerous other successful restorations highlight the model's strong capability in handling acronyms and shorthand notations.

The model also excels at extracting locations, values, and assertion statuses, even when attributes appear in distant sentences requiring inference. For instance: (1) In the anonymized sentence:

**"Broncoscopy on [DATE], [DATE], [DATE], [DATE], [DATE], [DATE]"**, the model correctly

infers the value *"6"* with the unit *"instances"*. (2) In the sentence: **"After this the pt. vomited\nfeculent material and then aspirated to the LLL with resultant\nhigh fever (105) and hypoxemia necessitating ICU observation overnight"**, the model successfully extracts *"hypoxemia"* despite irregular line breaks and assigns *"Left Lower Lobe"* as the location, inferring it from the abbreviation *"LLL"* while skipping over *"fever"*. (3) In the sentence: **"concerning for \npartial small bowel obstruction and/or ileus"**, the phrase *"small bowel obstruction"* is correctly assigned the status *"possible"*. (4) Furthermore, GENIE avoids overfitting allergy statuses to *"conditional"* and accurately assigns *"absent"* in the sentence: **"Allergies:\nPatient recorded as having No Known Allergies to Drugs"**.

Though we did not quantitatively measure the correctness of purpose, the ability of this attribute is good. The inference of purpose mainly comes from world knowledge of LLMs, and then combines with specific situations of the patient. For example, the purpose of a chest X-ray will be "**obtain detailed images of the chest**", which typically remains the same when this procedure is listed. The purpose of a medicine will more frequently be curing the patient's corresponding symptoms, such as "**treat thyroid hormone deficiency**" for "**levothyroxine**" and "**relieve constipation**" for "**colace**".

*Limitations and future work*

Despite these strengths, we identified some suboptimal cases. These include incorrect acronym restorations, attribute mismatches, and missed terms during Named Entity Recognition (NER). Examples of these cases are also included in the Appendix and will be discussed in the subsequent section.

Firstly, the extracted values and units are not consistently unified into a standard format. Variations such as acronyms (e.g., "mmHg" versus "millimeters of mercury"), scientific notation (e.g., "/cm³" versus "10³/cm³"), and differing range expressions (e.g., "<" versus "less than") can undermine direct downstream applications. This inconsistency largely stems from the variability in GPT outputs, where only a portion of non-standard expressions pass the evaluation process. Secondly, attributes like locations and purposes are primarily generated by GPT, making it impractical to verify the correctness of each training data entry. While errors are rare, they do occur—such as dislocation of attributes—which may adversely impact model performance. Additionally, ambiguity in acronyms sometimes leads to phrase extraction errors or incomplete restoration.

Based on these observations, we propose several improvements for future versions of the GENIE model. A passage-specific acronym table generated by GPT will be constructed to refine term extraction results, serving both the training stage (for deduplication) and the evaluation stage. This table may also be applied directly for post-processing of model outputs where feasible. Furthermore, cross-validation using other offline LLMs will be explored for verifying term attributes and standardizing outputs. However, it must be acknowledged that certain errors arise from inherent limitations of current LLMs and may be challenging to address fully.

Beyond the scope of the current work, we have also developed a Chinese EHR structuring model by translating MIMIC-III data and aligning the extracted results with their corresponding texts. This effort represents an initial step toward generalizing our method for EHR structuring tasks in low-resource languages.

## Conclusion

In this work, we introduce the **GENIE** model—*Generative Note Information Extraction*—to perform end-to-end structuring of electronic health record (EHR) data. Our model integrates named entity recognition, assertion status detection, and entity attribute extraction into a unified pipeline. This end-to-end approach eliminates the need for separate processing stages and demonstrates strong performance across all tasks. In future work, we aim to address occasional error patterns, further improve model robustness, and extend its applicability to multiple languages.